# AI-Driven Real-Time Monitoring of Ground-Nesting Birds: A Case Study on Curlew Detection Using YOLOv10

**Carl Chalmers** [1,*], **Paul Fergus** [1], **Serge Wich** [1], **Steven N Longmore** [1], **Naomi Davies Walsh** [1], **Lee Oliver** [2], **James Warrington** [2], **Julieanne Quinlan** [2] and **Katie Appleby** [2]

1. Liverpool John Moores University; p.fergus@ljmu.ac.uk, s.wich@ljmu.ac.uk, s.n.longmore@ljmu.ac.uk, n.j.walsh@2022.ljmu.ac.uk
2. Game & Wildlife Conservation Trust; loliver@gwct.org.uk, jwarrington@gwct.org.uk, jquinlan@gwct.org.uk, kappleby@gwct.org.uk
* Correspondence: c.chalmers@ljmu.ac.uk

**Abstract:** Effective monitoring of wildlife is critical for assessing biodiversity and ecosystem health, as declines in key species often signal significant environmental changes. Birds, particularly ground-nesting species, serve as important ecological indicators due to their sensitivity to environmental pressures. Camera traps have become indispensable tools for monitoring nesting bird populations, enabling data collection across diverse habitats. However, the manual processing and analysis of such data are resource-intensive, often delaying the delivery of actionable conservation insights. This study presents an AI-driven approach for real-time species detection, focusing on the curlew (*Numenius arquata*), a ground-nesting bird experiencing significant population declines. A custom-trained YOLOv10 model was developed to detect and classify curlews and their chicks using 3/4G-enabled cameras linked to the Conservation AI platform. The system processes camera trap data in real-time, significantly enhancing monitoring efficiency. Across 11 nesting sites in Wales, the model achieved high performance, with a sensitivity of 90.56%, specificity of 100%, and F1-score of 95.05% for curlew detections, and a sensitivity of 92.35%, specificity of 100%, and F1-score of 96.03% for curlew chick detections. These results demonstrate the capability of AI-driven monitoring systems to deliver accurate, timely data for biodiversity assessments, facilitating early conservation interventions and advancing the use of technology in ecological research.

## 1. Introduction

Ground-nesting birds are among the most vulnerable wildlife groups, facing an array of environmental pressures that threaten their survival [1]. These species, which include lapwings, skylarks, and curlews rely on open landscapes to breed, where their nests and chicks are particularly exposed to threats such as habitat loss and predation [2]. Agricultural intensification, urban development and changes in land use have significantly reduced the availability of suitable nesting sites, leaving many ground-nesting birds struggling to maintain stable populations [3]. Predation by meso-predators such as foxes, badgers, and corvids further exacerbate the challenge, often leading to high rates of egg and chick mortality [4]. Ground-nesting birds play an important ecological role in the maintenance of grassland and wetland ecosystems through the regulation of insect populations and the construction of ground nests that encourage diverse vegetation structures [5]. By feeding on insects and other invertebrates, they help control pest populations that could otherwise affect plant communities and agricultural crops [6]. Their nesting activities contribute to soil aeration and nutrient redistribution, fostering plant diversity and supporting a range of other wildlife species.

Conservation of these ground nesting birds is essential not only for their survival but also for the overall health and resilience of the ecosystems they inhabit.

Among these species, the curlew (*Numenius arquata*) is of particular concern, exemplifying the challenges faced by ground-nesting birds [7]. Once widespread across the UK, curlew populations have experienced dramatic declines in recent decades, driven by habitat degradation and predation [8]. This has left the species on the brink of regional extinction, with urgent conservation efforts needed to reverse this trend [9].

Efforts to protect vulnerable bird populations have largely focused on habitat restoration, predator control, and traditional nest monitoring [10]. While these methods have proven valuable, they face significant limitations that hinder their effectiveness. Traditional nest monitoring, which typically involves physical observation or manual review of historical camera trap data, is labor-intensive, time-consuming, and logistically challenging—particularly for species like curlews that nest in remote or hard-to-access areas [11].

The use of camera traps has offered a non-invasive solution for monitoring wildlife, enabling researchers to gather large datasets over extended periods and across multiple sites [12]. However, the manual collection and analysis of camera trap data remains a major bottleneck [13]. Reviewing the acquired images manually is resource-intensive, delaying the ability to implement timely conservation actions [14].

While AI-driven solutions exist to automatically analyse camera trap data, many focus on filtering out blank images and broadly categorising the presence or absence of animals, people and cars, rather than reporting species-level classifications [15]. This limitation is particularly problematic for biodiversity monitoring, where distinguishing between visually similar species is critical for targeted conservation efforts [16]. Solutions that offer species-level classification often fail to encompass common ground-nesting bird species, making them inadequate for monitoring these vulnerable populations effectively.

Furthermore, many solutions lack real-time data processing. Camera trap images are often stored on SD cards and retrieved manually, resulting in significant delays between data collection and analysis [17]. This workflow is impractical for species like curlews, where immediate interventions may be necessary to protect eggs or chicks from predation. Some AI systems, such as MegaDetector [18], PyTorch Wildlife [19], Wildlife insights [20] and INaturalist [21], have demonstrated potential in automating aspects of wildlife monitoring. However, their effectiveness is often constrained by reliance on pre-processed datasets, imbalanced class representation [22], inability to generalise across ecosystems, and a lack of integration with real-time data streams [23] [24].

Most existing systems prioritise reducing data volume rather than enabling proactive responses, making them ill-suited for dynamic conservation scenarios where timely decision-making is crucial [25]. To address these challenges, there is a need for AI-enabled solutions that combine high accuracy, species specific classification, and real-time processing capabilities. By automating the detection of vulnerable species like curlews and providing immediate insights to researchers, such systems can help overcome the limitations of traditional methods and existing AI approaches. These innovations have the potential to revolutionise the way we monitor and protect ground-nesting birds therefore enabling more efficient and effective conservation strategies.

In this study, we present an automated object detection and classification pipeline capable of identifying curlews and curlew chicks (Figure 1). Our model utilises a custom trained YOLOv10 Deep Learning (DL) model, which is integrated with the Conservation AI platform [26]. This model is capable of detecting and classifying curlews and their chicks received from real-time 3/4G-enables camera traps. The approach automates species detection and supports large-scale biodiversity surveys, to enable more timely and accurate ecological assessments, especially for vulnerable species like the curlew. By focusing on curlew monitoring, this study demonstrates the applicability of a scalable, efficient solution for tracking curlews over extended periods to identifying key opportunities for intervention. Automating the analysis of camera trap data in this way accelerates data processing while providing real-time alerts on ecologically significant events, such as nesting activity or chick presence, enabling further analysis for timely interventions to support curlew population recovery.

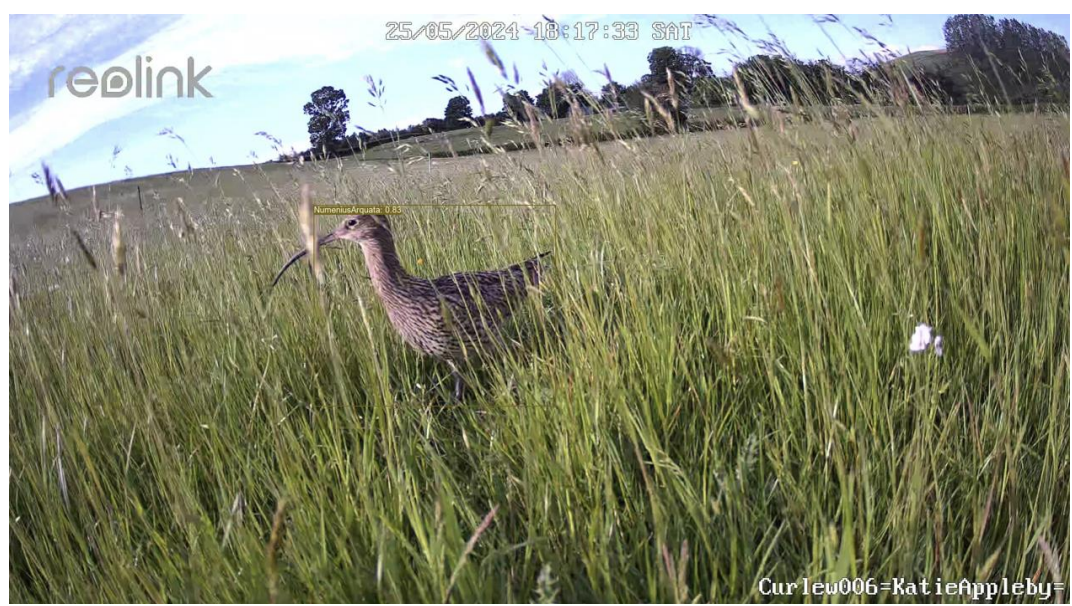

**Figure 1.** Camera trap image of a Curlew visiting one of areas used in the study.

The remainder of this paper is structured as follows: Section 2 outlines the methodology, including data collection, model training, and evaluation metrics. Section 3 presents the results, including both model performance and real-time inference for the curlew case study, followed by the discussion in Section 4. The conclusions and future directions are presented in Section 5.

**2. Materials and Methods**

In this section, we outline the dataset used in the study, detailing the modeling approach and evaluation metrics applied to assess the performance of the trained model. Additionally, we cover the data tagging and pre-processing stages, including the proposed methods and associated technologies employed to prepare the data for modelling.

*3.1. Data Collection and Description*

The image dataset contains images of 26 distinct species and objects found in the UK: *Person, Vulpes vulpes (Red fox), Dama dama (European fallow deer), Capreolus capreolus (Roe Deer), Erinaceus europaeus (European hedgehog), Capercaillie cock (Wood grouse), Capercaillie hen (Wood grouse), Bos taurus (Cattle), Canis familiaris (Domestic dog), Cervus elaphus (Red deer), Oryctolagus cuniculus (European rabbit) Meles meles (European badger), Buteo buteo (Common buzzard), Accipiter gentilis (Northan goshawk), Felis catus (Domestic cat), Sciurus carolinensis (Eastern grey squirrel), Sciurus vulgaris (Red squirrel), Martes martes (European pine martin), Phasianus colchicus (Common pheasant), Passer domesticus (House sparrow), Ovis aries (Domestic sheep), Columba palumbus (Common wood pigeon), Numenius arquata (Common curlew), Numenius arquata chick (Common curlew), Capra hircus (Domestic goat) and Calibration pole (Calibration pole)* which were obtained through various conservation partners and private camera deployments.

The dataset used in this study comprised a total of 38,740 image files. The average resolution across the dataset was 972 x 769 pixels, aligning with the typical input resolution supported by most camera trap systems. An analysis of the resolution distribution revealed no significant outliers that would adversely affect model training as shown in Figure 2. As a result, no images were excluded prior to tagging. Conducting this input resolution analysis was essential for determining the appropriate aspect ratio coefficient, which was then incorporated into the hyperparameter configuration for training the model.

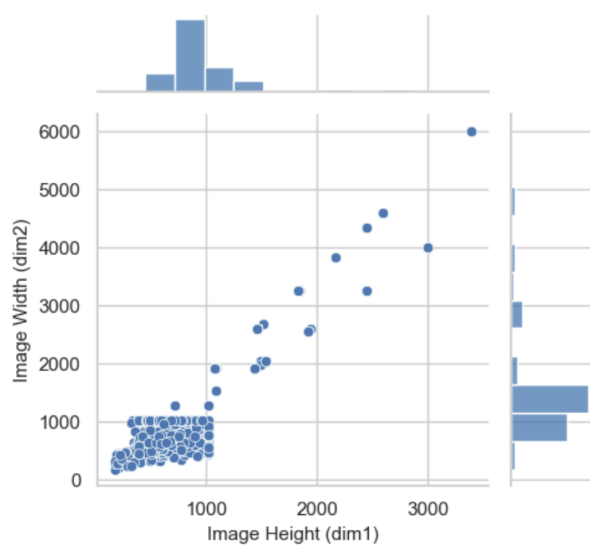

**Figure 2.** Distribution of image resolutions across the dataset, highlighting any outliers that may impact model training and performance.

*2.2. Data Pre-Processing*

The data tagging process was conducted using the Conservation AI tagging platform, where bounding boxes were applied to delineate regions of interest within each image. These tagged regions were exported in Extensible Markup Language (XML) format, adhering to the Pascal VOC standard. Images deemed to be of poor quality or unsuitable for model training were labeled as "no good" and systematically excluded from the final dataset. The number of images per object ranged from approximately 1,100 to 2,500, resulting in a class imbalance (Figure 3) due to reduced representation in the original dataset. In total, 38,740 objects were tagged across the dataset.

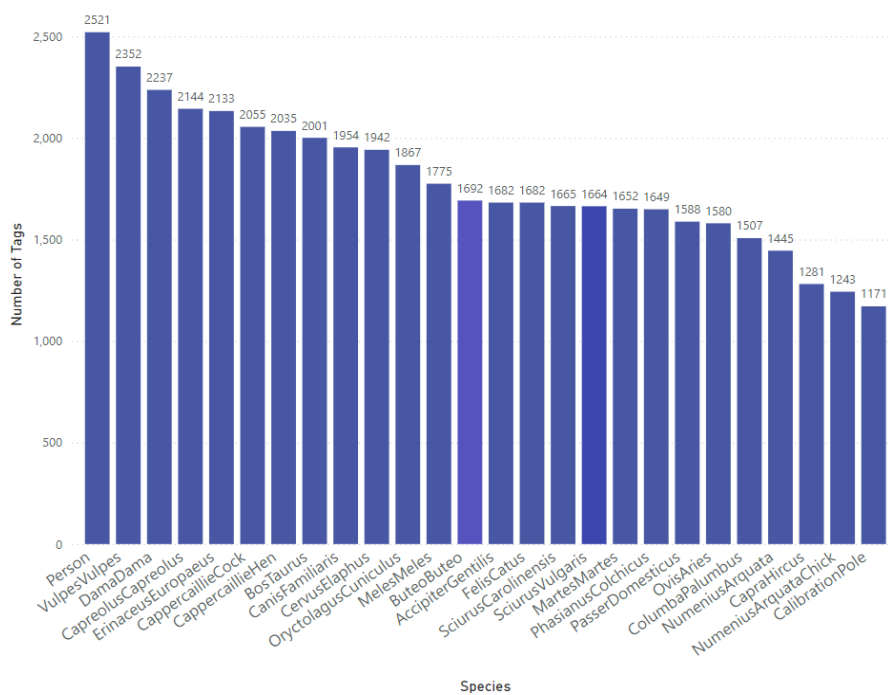

**Figure 3.** Class distribution of annotated data showing the species name on the x axis and the number of tags per species on the y axis.

The labeled data was converted to the YOLO annotation format using a Python script. The dataset, consisting of images and corresponding labels, was randomly split into 80% for training, 10% for validation, and 10% for testing, based on the tagged annotations. The XML files were converted directly into YOLO-specific text files, where each object is represented by its class and bounding box coordinates.

*2.3. Model Selection*

The YOLOv10x model was employed for both object detection and classification across 26 different species and objects. The model features 29.5 million parameters, balancing high model capacity with computational efficiency, which allows it to perform complex detection tasks at scale without excessive hardware requirements. Unlike the two-stage approach used in models such as Faster-RCNN, YOLOv10x integrates detection and classification in a single-stage process, significantly improving both speed and efficiency. This model utilises the CSPDarknet backbone, which extracts essential feature maps, and incorporates a Path Aggregation Network (PAN) for feature fusion, improving detection of objects at multiple scales. YOLOv10x further refines object detection with its anchor-free detection mechanism, eliminating the need for predefined anchor boxes, and leverages dynamic convolution to enhance classification accuracy [27]. By predicting bounding boxes and class probabilities simultaneously, the model achieves high-speed inference while maintaining high accuracy, making it well-suited for real-time applications in ecological monitoring. These optimisations ensure that YOLOv10x performs well even in complex tasks involving multiple classes, such as wildlife monitoring for biodiversity assessments. Figure 4 shows the YOLOv10 architecture.

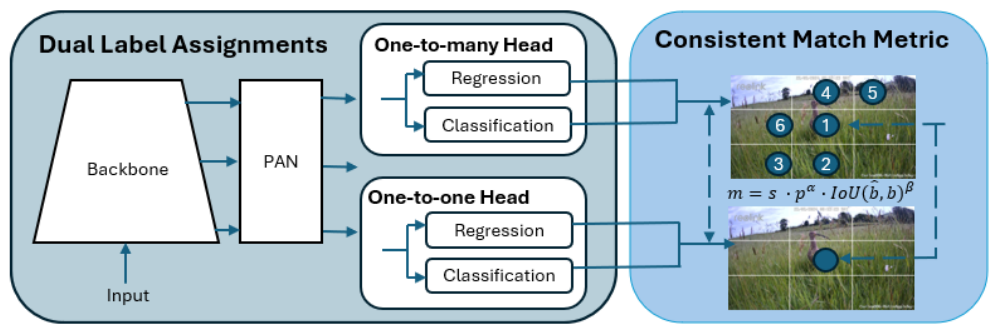

**Figure 4.** YOLOv10 architecture used in the study.

*2.4. Transfer Learning*

Transfer learning enables the adaptation of a pre-trained model to new tasks by fine-tuning its learned parameters for novel objects or species of interest [28]. This technique is critical when working with smaller datasets, as training DL models from scratch on limited data can lead to poor performance due to reduced feature representation and low variance [29]. By leveraging a model pre-trained on a much larger dataset, transfer learning helps mitigate these challenges, reducing the data required for effective training while maintaining robust accuracy. This makes it particularly valuable for applications with limited resources, such as ecological monitoring and biodiversity assessments.

In this study, YOLOv10x was employed as the foundation model for transfer learning. The model was initially pre-trained on the MS COCO dataset, a large object detection dataset allowing it to learn generalised features across a wide variety of objects. The pre-training of YOLOv10x was conducted using 8 NVIDIA 3090 GPUs over a span of approximately 120 hours (500 epochs), as described in the original paper [30]. This setup enabled efficient optimisation of the model parameters, creating a strong foundation for transfer learning.

By fine-tuning YOLOv10x on a smaller dataset of 26 species, the model's learned weights were adapted to recognise new species with high accuracy, despite the limited size of the training data. This approach makes it feasible for a wide range of users to deploy high-performance models without requiring extensive computational resources or large datasets, making it an accessible solution for conservation and ecological studies.

*2.5. Modelling*

Model training was conducted on a custom-built Gigabyte server equipped with an AMD EPYC 7252 series processor and 128 GB of RAM. To support intensive computational tasks, the server was fitted with an additional GPU stack comprising 8 Nvidia Quadro A6000 graphics cards, which provided a combined total of 384 GB of GDDR6 memory. The training environment was set up using PyTorch 2.0.1 and CUDA 11.8, forming the core software components of the training pipeline. The following key hyperparameters were used during the training process:

- The image size is set to 640 pixels, ensuring the balance between detection accuracy and computational efficiency for high-resolution input. This resolution closely aligns with the mean resolution of the acquired dataset.
- The batch size coefficient is set to 256, allowing for a stable weight update process without exceeding the memory limitations of the available GPU hardware.
- The epochs parameter is configured to 50, ensuring adequate time for convergence while preventing overfitting in the model.
- The learning rate is set to 0.01, providing a balanced update speed, which prevents rapid shifts in response to errors.
- The momentum is set to 0.937, improving the training stability by maintaining model direction toward the minima during gradient descent.

To enhance the generalisation of the model and reduce overfitting, several augmentation techniques were applied during training. It's important to note that these augmentations did not increase the size of the dataset, as the images were randomly sampled and modified in real-time during training. The applied augmentations included the following:

- Hue adjustment (hsv_h=0.015): The hue of images was randomly adjusted by up to 1.5%, introducing slight color variations.
- Saturation adjustment (hsv_s=0.7): Saturation levels were altered by up to 70%, providing variety in color intensity.
- Brightness adjustment (hsv_v=0.4): The brightness (value) was adjusted by up to 40%, simulating different lighting conditions.
- Horizontal flip (fliplr=0.5): Images were horizontally flipped with a 50% probability, increasing the model's invariance to directionality.
- Translation (translate=0.1): Images were randomly shifted by up to 10%, helping the model handle variations in object positioning.
- Scaling (scale=0.5): The size of objects in the images was adjusted by scaling up to 50%, improving detection at different object sizes.
- Random erasing (erasing=0.4): Applied to 40% of the images, simulating partial occlusions by randomly removing parts of the image.

*2.6. Model Inferencing*

The trained model was frozen and exported in ONNX format to facilitate inferencing. It was deployed and hosted via a publicly accessible website (www.conservationai.co.uk), developed by the authors for real-time species detection. The inference process was conducted on a custom-built server equipped with an Intel Xeon E5-1630v3 CPU, 256 GB of RAM, and an NVIDIA Quadro RTX 8000 GPU. The software stack for model inferencing utilized NVIDIA Triton Inference Server (version 22.08) [30], running within a Docker environment on Windows Subsystem for Linux 2 (WSL2). Given the high computational capabilities of the NVIDIA RTX 8000 GPU, no additional model optimisation techniques, such as quantisation or pruning, were necessary to enhance performance, ensuring efficient and accurate inferencing without compromising model complexity.

3/4G cellular cameras were deployed at 11 locations across the UK, as depicted in Figure 5. The cameras were configured to capture images at a resolution of 1920 x 1072 pixels with a 96 DPI (dots per inch), closely matching the resolution of both the training images and the image size used during model training. The infrared (IR) sensor sensitivity was set to medium, and when triggered, the acquired images were automatically uploaded to the platform for classification using the Simple Mail Transfer Protocol (SMTP). Each camera was powered by a lithium battery (7800mAh Li9) and recharged via solar panels throughout the duration of the study. Figure 5 illustrates one of the cameras used in this deployment.

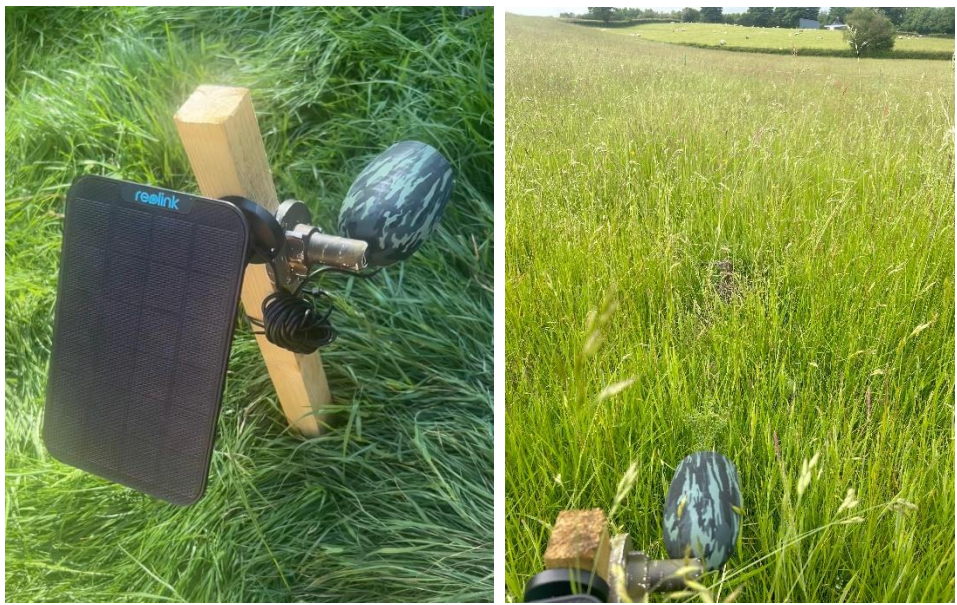

**Figure 5.** 3/4G camera trap used for real-time inference. Throughout the study the camera trap was continuously charged using a solar panel.

The end-to-end inferencing pipeline as shown in Figure 6 begins with data capture from the camera and concludes with the public-facing Conservation AI site highlighted in Figure 7. The system is designed to interface with a wide variety of cameras for real-time inference using standard protocols. When the IR sensor is triggered, the camera automatically transmits the image and associated meta data to a dedicated Simple Mail Transfer Protocol (SMTP) server running on the Conservation AI platform. The acquired data is automatically classified via the Triton Server REST API and saved to an internal database. In cases where field communication is unavailable, image files can be batch uploaded through the website, desktop client, or REST API for offline inferencing, ensuring flexibility and scalability in data processing.

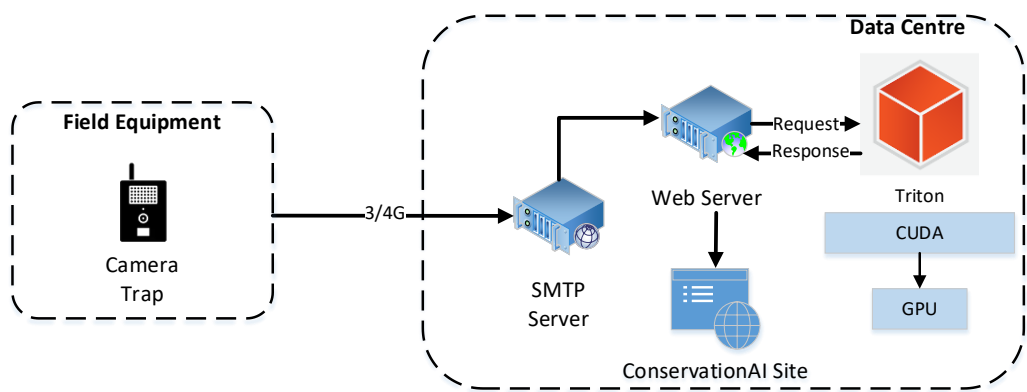

**Figure 6.** End-to-end inferencing pipeline for the Conservation AI platform.

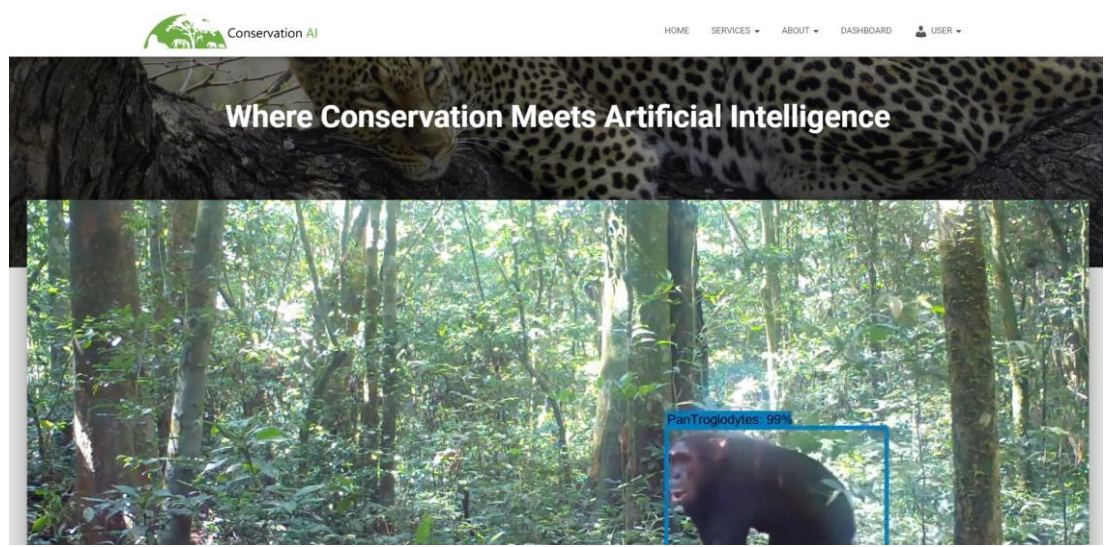

**Figure 7.** Conservation AI inferencing platform using Triton server.

The end-to-end inferencing pipeline as shown in Figure 6 begins with data capture from the camera and concludes with the public-facing Conservation AI site highlighted in Figure 7. The system is designed to interface with a wide variety of cameras for real-time inference using standard protocols. When the IR sensor is triggered, the camera automatically transmits the image and associated meta data to a dedicated Simple Mail Transfer Protocol (SMTP) server running on the Conservation AI platform. The acquired data is automatically classified via the Triton Server REST API and saved to an internal database. In cases where field communication is unavailable, image files can be batch uploaded using the Conservation AI desktop application, or programmatically using the REST API for offline inferencing, ensuring flexibility and scalability in data processing.

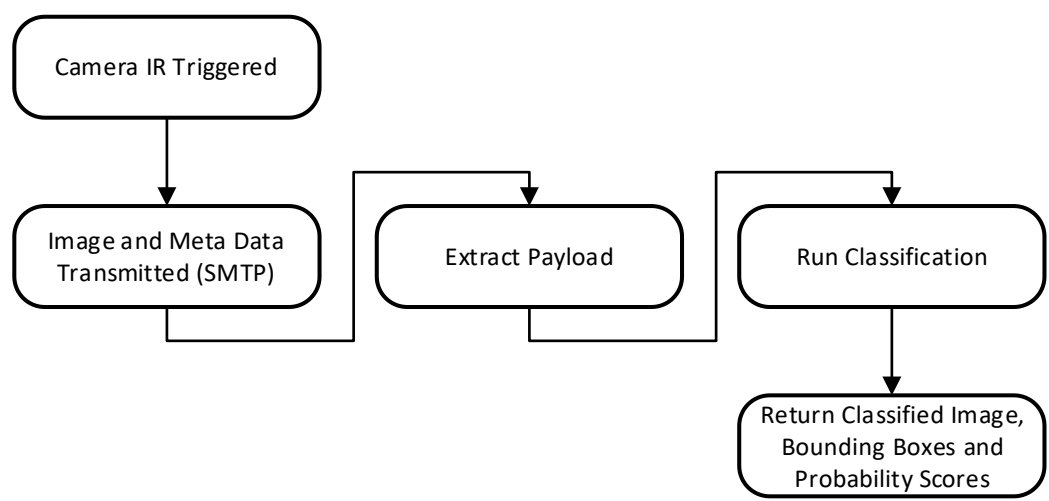

**Figure 8.** Inference processing pipeline.

*2.7. Evaluation Metrics Training*

The model is evaluated using the test split following training to assess its generalisation performance before deployment in real-time systems. This evaluation provides a comprehensive view of the model's behavior on unseen data. To measure performance, several key metrics were employed: Precision, Recall (Sensitivity), F-1 Score, and Accuracy, along with visualisations such as the Precision-Confidence Curve, Recall-Confidence Curve, and F-1 Confidence Curve.

The precision-confidence curve provides a visual representation of how the model's precision—its ability to make accurate positive detections—varies across different confidence thresholds. This curve helps in assessing the model's effectiveness in minimising false positives as the confidence level increases. A higher precision at higher confidence thresholds indicates that the model makes more accurate predictions when it is more certain about its detections.

The recall-confidence curve illustrates how the model's ability to correctly detect true positives changes across varying confidence thresholds. This curve is essential for understanding the model's sensitivity to false negatives. By examining the recall at different confidence levels, the curve provides insights into how well the model can capture all relevant objects, even as the confidence threshold increases.

The F-1 confidence curve balances precision and recall at different confidence levels. This curve provides a more holistic view of the model's performance, as the F-1 score is a harmonic mean of both metrics. A consistently high F-1 score across confidence thresholds indicates that the model is both accurate and reliable in detecting and classifying objects, with minimal trade-off between precision and recall.

*3.8. Evaluation Metrics Inference*

The performance of the trained model during inference is evaluated by analysing images transmitted from real-time cameras during the trial. Post-training analysis is crucial, as model performance can vary significantly when deployed for real-world tasks due to variance not captured in the training data. Precision, Sensitivity (Recall), Specificity, F1 Score, and Accuracy were employed to assess the model's performance on inference data. These metrics were derived from True Positives (TP), False Positives (FP), True Negatives (TN), and False Negatives (FN). Precision, for instance, is defined as:

$$Precision = \frac{TP}{TP + FP} \tag{7}$$

Precision measures the proportion of true positive detections among all positive predictions made by the model. In the context of this study, it indicates how often the animal detected and classified by the model matches the ground truth, reflecting its accuracy in avoiding misclassifications. Recall, on the other hand, is defined as:

$$Recall = \frac{TP}{TP + FN} \tag{8}$$

Recall measures the proportion of true positives correctly identified by the model. In the context of object detection, it evaluates how effectively the model detects and correctly matches the ground truth labels. Recall also helps identify the number of false negatives (instances the model missed). The F1 Score, which balances Precision and Recall, is defined as:

$$F - 1\ Score = 2 * \frac{Precision * Sensitivity\ (Recall)}{Precision + Sensitivity\ (Recall)} \tag{9}$$

The F1 Score represents the harmonic mean of Precision and Recall, providing a single metric that balances both. A high F1 Score indicates that the model achieves a good trade-off between Precision and Recall. In the context of object detection, high Recall signifies the model's ability to detect and localise most objects in an image, while Precision ensures that the detections are accurate. Lastly, Accuracy is defined as:

$$Accuracy = \frac{TP + TN}{TP + TN + FP + FN} \tag{10}$$

Accuracy provides an overall evaluation of the model's performance in detecting and correctly classifying objects within an image. However, its relevance diminishes in the context of unbalanced datasets, as it can be misleading when certain classes are over- or under-represented. Therefore, accuracy should always be considered alongside other metrics, such as Precision, Recall, and the F1 Score, to provide a more comprehensive assessment of model performance.

## 3. Results

We present the results in two parts: firstly, the training of the YOLOv10x model using the hyperparameters outlined in the methodology section; secondly, the inference results which demonstrate the model's performance in a real-world setting, using the 3/4G real-time cameras. Please note that the term classes represent the individual species/objects and therefore the terms species and classes are used interchangeably in this paper.

### 3.1. Training Results for UK Mammals Model

Using the tagged dataset, the data was randomly split into a training set, comprising 80% of the data, a validation set of 10% to train and fine tune the model and a test set of 10% to ascertain the model's performance on unseen data. The model was trained over 50 epochs using a batch size of 256 to determine the best fit. During training, there was no overlap between the training and validation loss, indicating that no overfitting occurred.

The Precision-Recall (PR) curve presented in Figure 9 provides a detailed assessment of the model's performance across all 26 classes. The model achieved a mean Average Precision (mAP) of 0.976 at a 0.5 Intersection over Union (IoU) threshold, indicating a high level of overall detection accuracy. This high mAP value reflects the model's strong capability in both detecting and correctly classifying species across diverse categories. The curve demonstrates that the model sustains high precision even as recall increases, which is a clear indicator of the model's robustness and reliability in minimising false positives while capturing true positives. The shape of the PR curve remains stable across most classes, suggesting that the model is well-calibrated for a wide range of detection tasks. However, some deviations are observed in the curves for individual classes, indicating that there are certain species for which the model's performance could be improved. These deviations could be attributed to insufficient training data for specific classes or challenges in distinguishing between visually similar species. Addressing these issues through further fine-tuning or the inclusion of additional training data may enhance the model's ability to generalise across all species.

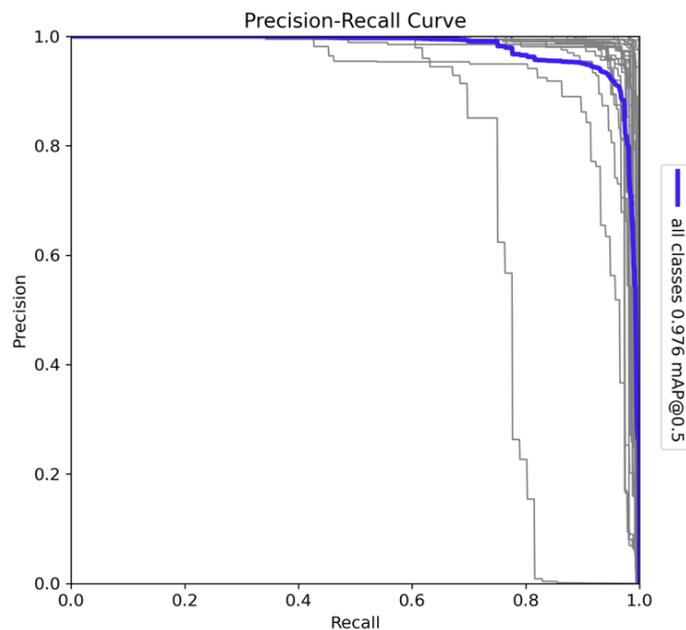

**Figure 9.** Precision-Recall (PR) curve for the model trained on UK species and objects.

The Precision-Confidence curve for the model, as illustrated in Figure 10, provides a detailed analysis of the model's reliability across all 26 classes. The steep ascent of the curve indicates that the model achieves high precision even at relatively low confidence thresholds. This suggests that the model can make accurate predictions with moderate confidence, maintaining strong performance across a broad range of confidence levels.

At the upper end of the confidence spectrum, the model achieves perfect precision at a confidence level of 1.0 (with a precision value of 1.00 at 0.996). This outcome demonstrates that the model is consistently accurate when it is highly confident in its predictions. Minor variations observed in the curves for individual classes reflect some inherent classification challenges, likely due to interspecies similarities or variance in visual features. However, the overall correlation between precision and confidence confirms the robustness of the model in making reliable detections, even when confidence levels are not maximised. This consistent performance across confidence levels makes the YOLOv10x model highly suitable for real-world deployment, as it can confidently handle predictions with both low and high certainty, maintaining accuracy and minimising false positives across varying conditions. These characteristics underscore the model's effectiveness in tasks requiring precision, such as wildlife species classification and biodiversity monitoring.

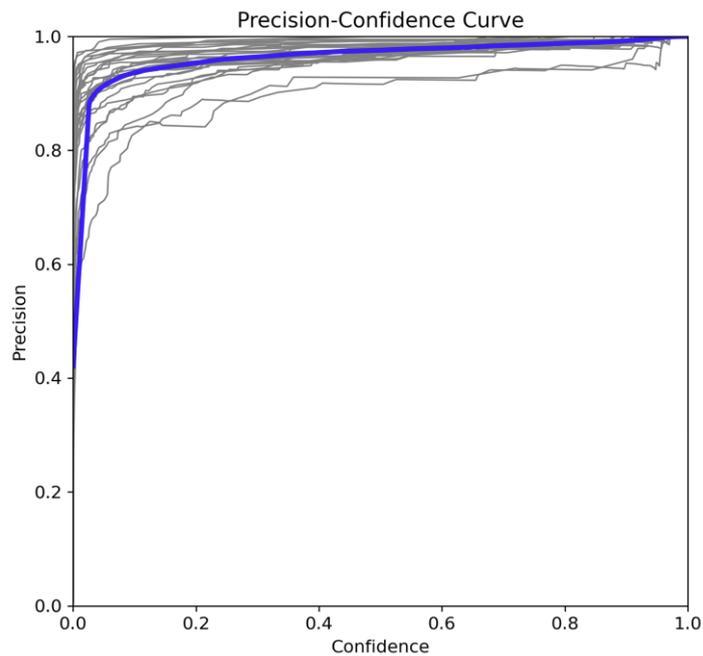

**Figure 10.** Precision-Confidence curve for the model trained on UK species and objects.

The Recall-Confidence curve, as presented in Figure 11, illustrates the relationship between recall and confidence across all 26 classes. At lower confidence thresholds, the model achieves near-perfect recall, with a recall value of 0.98 at a confidence threshold of 0.0. This indicates that the model is capable of capturing nearly all true positives when the confidence requirement is minimal, ensuring comprehensive detection coverage. As the confidence threshold increases, recall decreases gradually, with a more pronounced drop occurring near the highest confidence levels. This behavior reflects the trade-off between recall and precision: while the model becomes more confident in its predictions at higher thresholds, it detects fewer objects, thus reducing recall. The sharp decline in recall at higher confidence thresholds suggests that fewer positive instances are identified when the model prioritises high certainty, potentially missing some true positives. The variation in individual class curves indicates that certain species may require lower confidence thresholds to maintain high recall. This underscores the need for careful tuning of the confidence threshold depending on the class or task. Overall, the model exhibits strong recall performance at lower confidence levels, which ensures that it captures a wide range of objects, making it suitable for applications where detecting all possible instances is critical.

The F1-Confidence curve in Figure 12 offers a detailed analysis of the model's trade-off between precision and recall across varying confidence thresholds. The model achieves its peak F1 score of 0.96 at a confidence threshold of 0.387, which indicates that the model performs optimally at this threshold, striking an ideal balance between precision and recall. This high F1 score across a range of confidence levels demonstrates the model's robustness in balancing these two key metrics for object detection tasks. As the confidence threshold increases towards 1.0, there is a noticeable decline in the F1 score. This sharp drop suggests that, while the model becomes more confident in its predictions, it begins to miss more true positives, leading to a decrease in recall and a consequent decline in the F1 score. This behavior reflects the inherent trade-off between precision and recall: as precision increases, fewer true positives are detected, leading to a drop in overall performance as measured by the F1 score. Despite this, the model maintains a consistently high F1 score across most confidence levels, underscoring its suitability for tasks that require a delicate balance between precision and recall. The model's performance at these varying confidence thresholds indicates its effectiveness in handling complex detection scenarios, making it particularly well-suited for applications in species detection and classification within ecological monitoring frameworks.

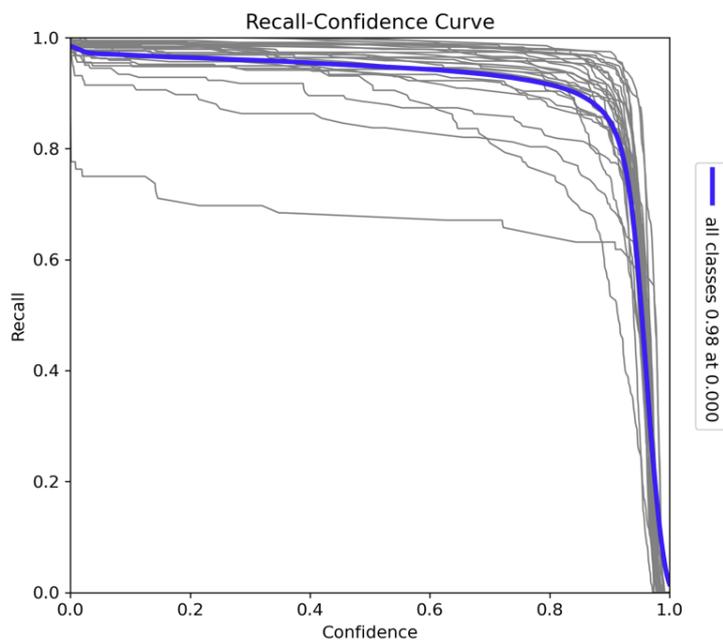

**Figure 11.** Recall-Confidence curve for the model trained on UK species and objects.

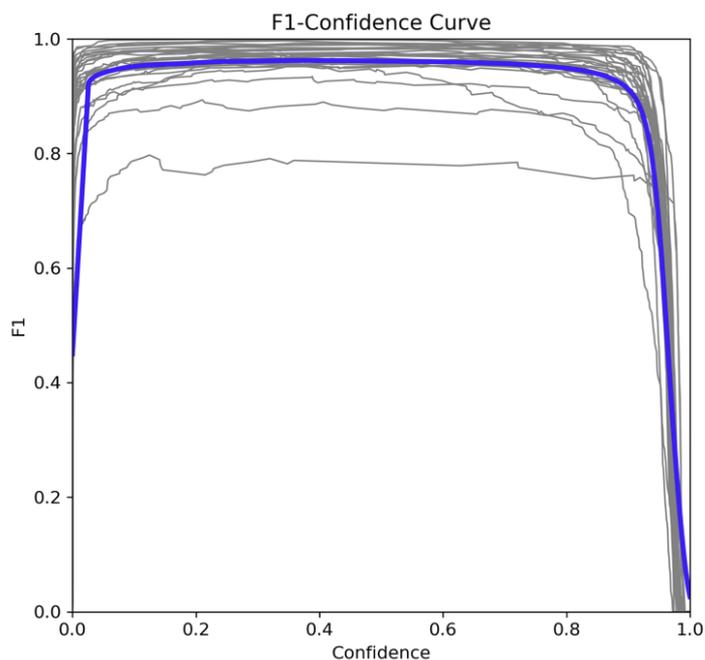

**Figure 12.** F1-Confidence curve for the model trained on UK species and objects.

The confusion matrix in Figure 13 provides a comprehensive assessment of the model's classification performance across the 26 species, vehicles, human subjects, and background categories used in this study. Each row corresponds to the predicted class, while each column represents the true class. The strong diagonal line seen in the matrix reflects that the majority of predictions are correct, with high values along the diagonal indicating accurate classification for most species. Notably, the most frequent species in the dataset, such as European hedgehog, Eastern Grey Squirrel, and Roe deer, exhibit the highest intensities along the diagonal, suggesting that the model has effectively learned to identify these species with high precision. The matrix also demonstrates the model's ability to distinguish between visually distinct species. However, several off-diagonal cells with lighter shading indicate instances of misclassification. These misclassifications, although relatively infrequent, occur between species that may share similar visual traits or appear in complex environmental conditions. For example, there are occasional confusions between roe deer and red deer, likely due to similarities in their appearance or habitat. Similarly, the misclassification between House sparrow and other bird species could be attributed to the challenge of distinguishing between small birds in various environmental conditions. Overall, the confusion matrix illustrates the model's strong classification performance, with high accuracy across the majority of classes and minimal instances of misclassification. The results highlight the model's robustness in distinguishing between a diverse range of species, supporting its applicability in ecological monitoring and species classification tasks.

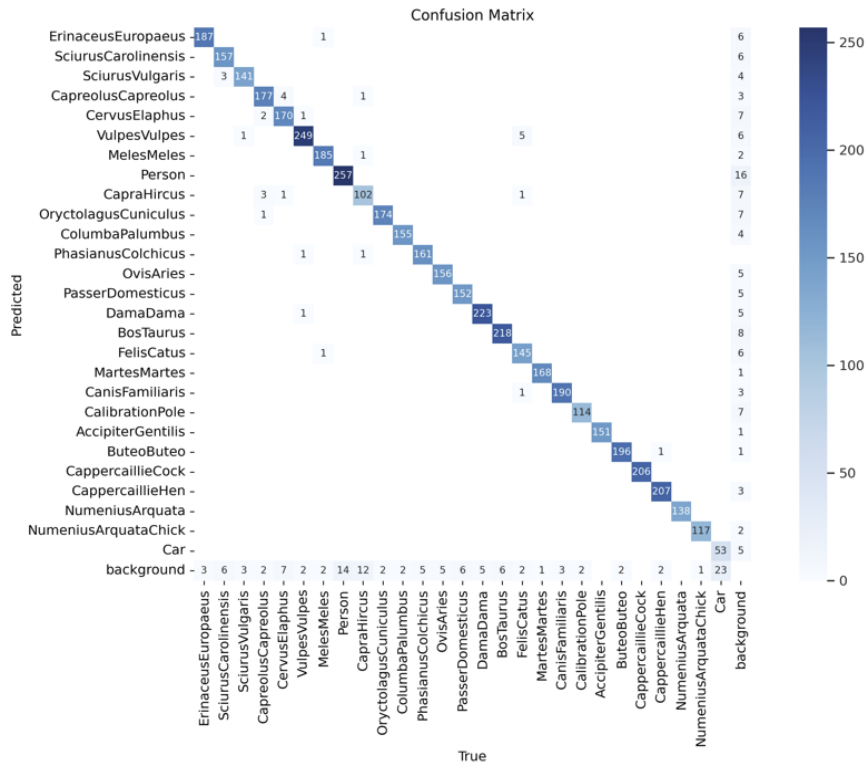

**Figure 13.** Confusion matrix for the model trained on UK species and objects.

*3.2. Model Deployment*

The trained model was deployed to evaluate its performance during the trial period. Using the inferencing pipeline and camera setup described in the methodology section, the system was deployed to monitor nesting curlews at 11 locations in Wales, UK. This deployment spanned from 20/05/2024 to 30/06/2024, during which a total of 1072 images were analysed by the platform. These images contained three distinct detections: domestic sheep (detected as: *Ovis aries*), Common curlew (detected as: *Numenius arquata*) and Common curlew (chick) (detected as: *Numenius arquata chick*).

14The detected objects in these images were evaluated to assess the model's detection accuracy and classification performance for curlews and their chicks, as well as for other species detected during the study. This analysis was critical for understanding the model's effectiveness in real-world scenarios and its suitability for long-term monitoring of species.

3.2.1. Performance Evaluation Results for Inference

The model achieved a high level of accuracy across most classes, with individual class accuracies ranging from 93.41% to 100% and an overall accuracy of 91.23% (Table 1). However, the performance metrics exhibited considerable variation among classes. Precision ranged from 0% to 100%, with Common pheasant (detected as: *Phasianus colchicus*) having the lowest value due to misclassifications and the absence of true instances in the dataset, while Common curlew (adult) and domestic sheep achieved perfect scores of 100%. The average precision of 60.34% suggests that while the model performs well for certain classes, there are significant issues with others.

Notably, some classes were discontinued from the analysis due to the absence of true instances observed during the study. Specifically, Common pheasant were not present in the dataset; however, some Common curlew (adult) were misclassified as Common pheasant. This misclassification led to false positives for Common pheasant, resulting in a precision and F1-score of 0% for this class. The sensitivity of the model was high across classes with actual instances, ranging from 90.56% for Common curlew (adult) to 100% for domestic sheep. The average sensitivity of 95.48% indicates that the model can correctly identify a large proportion of actual positive cases. Specificity was also notably high, with values between 96.64% and 100%. The species with the lowest specificity was Common pheasant, reflecting the misclassification issues, while Common curlew (adult) and domestic sheep each attained a specificity of 100%. The model's average specificity of 98.17% demonstrates a strong ability to correctly identify negative cases. The F1-scores varied widely across classes, from 0% for Common pheasant to 100% for domestic sheep. The species with the highest F1-score was domestic sheep (100%), while the lowest was Common pheasant (0%). The average F1-score across all classes was 58.88%, indicating that while the model performs exceptionally for some classes, it underperforms for others. The particularly low F1-score for Common pheasant underscores the impact of misclassification and the presence of classes without true instances in the dataset. This highlights the need for careful dataset curation and potential refinement of the model to address misclassification issues.

**Table 1.** Inference Performance Metrics for Camera Deployment.

|  | Accuracy | Precision | Sensitivity | Specificity | F1-Score |
|---|---|---|---|---|---|
| *Numenius arquata* | 93.41% | 100% | 90.56% | 100% | 95.05% |
| *Numenius arquata chick* | 97.51% | 100% | 92.35% | 100% | 96.03% |
| *Ovis aries* | 100% | 100% | 100% | 100% | 100% |

3.2.2. Confusion Matrix for Inference Data

Sources of confusion among the inference data (Table 1) largely align with the performance metrics presented in Table 2. The model correctly predicts most of the classes with minimal error; the Common curlew (adult) and domestic sheep classes performed the best. However, misclassifications were observed, notably with some instances of Common curlew (adult) being incorrectly classified as Common pheasant, despite the latter not being present in the actual dataset. This misclassification suggests that the model may rely on features common between these species, leading to false positives. These results highlight weaknesses in the model's ability to accurately detect and classify certain species, which will need to be addressed in future training iterations. Improvements may include refining the training dataset, enhancing feature selection, and adjusting augmentation techniques to reduce misclassification rates.

**Table 2.** Confusion Matrix for Inference. The diagonal number indicates the TP for each of the classes.

| Actual \ Predicted | Numenius arquata | Numenius arquata chick | Phasianus colchicus | Ovis aries | Total Actual |
|---|---|---|---|---|---|
| Numenius arquata | 662 | 0 | 36 | 33 | 731 |
| Numenius arquata chick | 0 | 302 | 0 | 25 | 327 |
| Ovis aries | 0 | 0 | 0 | 13 | 13 |
| Total Predicted | 662 | 302 | 49 | 58 | 1072 |

**4. Discussion**

This study demonstrates the effectiveness of an AI-driven approach for real-time monitoring of ground-nesting birds, with a focus on curlew (*Numenius arquata*) detection. Leveraging the YOLOv10x architecture, the model was trained on a diverse dataset achieving high classification performance for curlews and their chicks. Integration with the Conservation AI platform allowed the model to be deployed in conservation study pipelines to process data in real-time and to deliver actionable insights.

During the evaluation phase, the model processed 1,072 images from 11 camera sites in Wales. The results showed an overall accuracy of 91.23%, with specificity reaching 98.17% and sensitivity at 95.48%. For curlew detections, the model achieved a precision score of 100%, reflecting its ability to accurately identify true positives without misclassifications. Similarly, the F1-scores for adult curlews and chicks was 95.05% and 96.03%, respectively, underscoring the model's robust performance in both precision and recall.

The model's performance remained consistent across diverse and challenging environmental conditions. Figures 14 and 15 showcase successful detections of adult curlews and chicks, even in complex backgrounds. This is particularly significant for smaller species like curlew chicks, which are prone to blending into their natural surroundings.

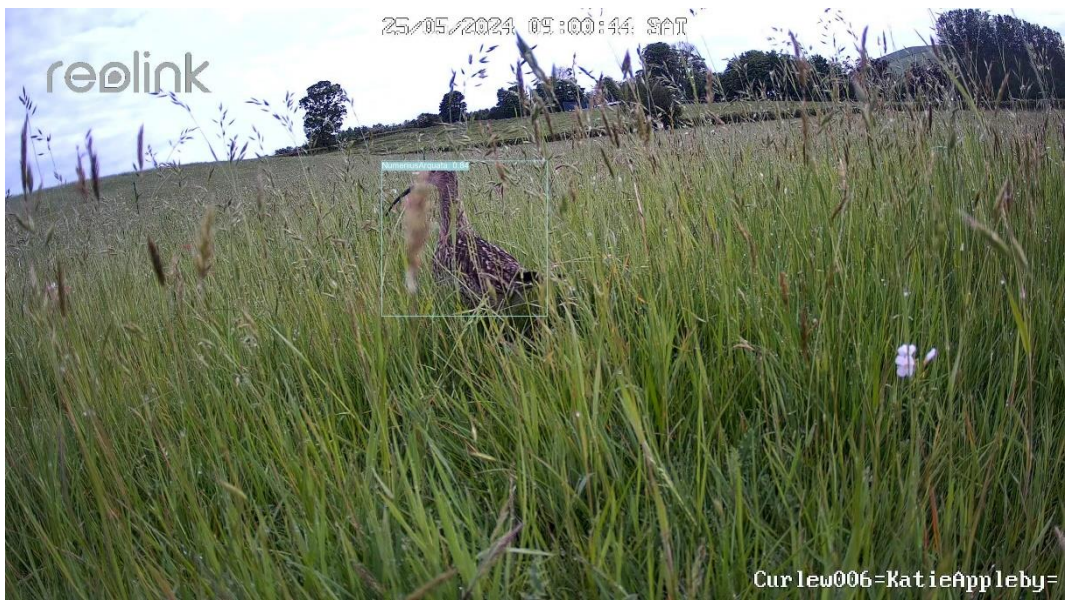

**Figure 14.** Example detections of Common curlew (adult) during the inference trial using one of the 3/4G cameras.

As illustrated in Figure 16, the model effectively distinguished curlews from dense vegetation, shadows, and other environmental noise, demonstrating its ability to generalise across variable conditions. This ability is critically important given the nature of the nesting locations of curlews.

The integration of the YOLOv10x model into a real-time monitoring system addresses critical limitations in traditional conservation methods. By filtering out irrelevant images with an accuracy of 98.28%, the system significantly reduces the data processing burden on conservationists, enabling them to focus on relevant detections and respond more promptly to potential threats. Additionally, the automated detection pipeline mitigates delays inherent in manual data processing, providing an efficient and scalable solution for biodiversity monitoring. This enhanced responsiveness allows conservationists and researchers to intervene much quicker when alerts are raised, maximising the protection of curlew eggs and chicks against predation and other threats.

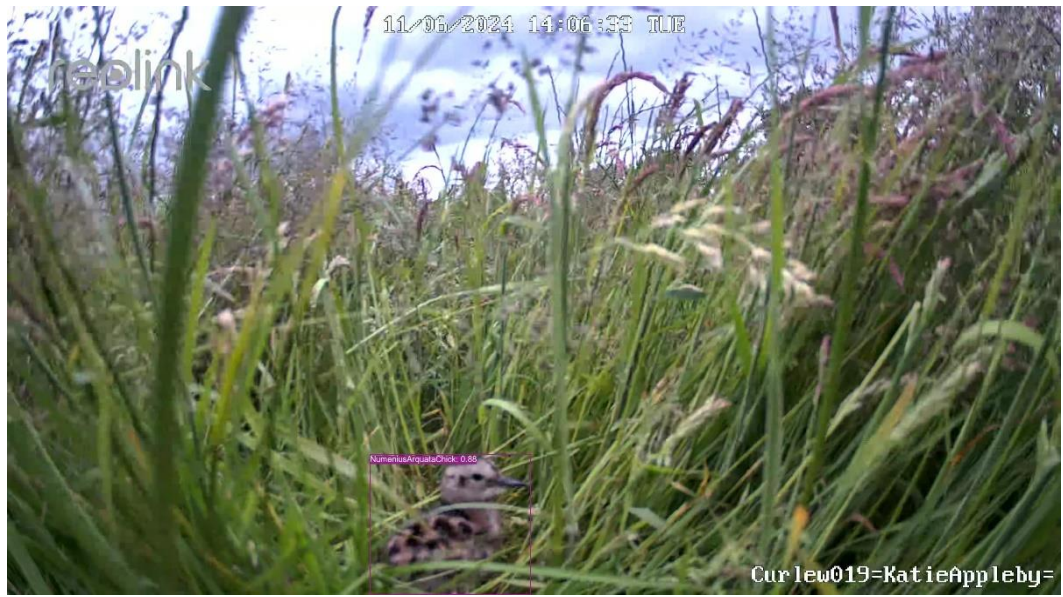

**Figure 15.** Example detection of a Common curlew chick during the trial using one of the 3/4G cameras.

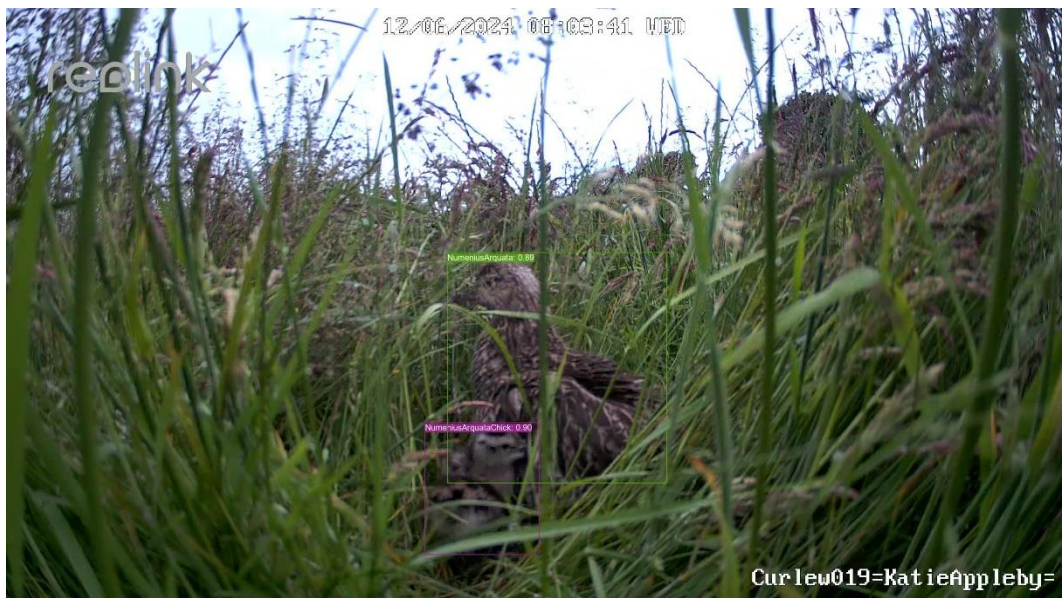

**Figure 16.** Example detection of a curlew and curlew chick obtained from a real-time camera.

Despite these achievements, several issues were identified. Misclassifications, particularly between common pheasants and curlews were notable during the study. Furthermore, correct camera placement was a critical factor for detecting smaller subjects like curlew chicks. Camera traps need to be deployed closer to nesting sites to ensure optimal trigger settings and to capture high-quality images in dynamic or cluttered environments, like the grassy locations in which curlews' nest. Not all camera trap installations in the study adhered to these guidelines, consequently some misdetections were observed.

**5. Conclusions**

This study presented an innovative AI-driven classification system designed to detect and monitor curlew populations using artificial intelligence and real-time 3/4G-enabled cameras. By addressing the significant challenges of large-scale monitoring in remote or hard-to-access areas, this system reduced the dependence on manual image collection and classification, tackling key bottlenecks in traditional biodiversity monitoring. The automated approach accelerates data processing and delivers timely, actionable information, enabling conservationists to intervene quickly which is an essential capability for species like the curlew, who are threatened by habitat loss and predation.

Using the YOLOv10x architecture and transfer learning, the model was deployed across 11 sites in Wales to evaluate its effectiveness in monitoring curlews and their chicks. Real-time data transmission facilitated immediate classification and documentation of curlew activity, showcasing the system's potential for proactive conservation efforts. For instance, this system could support breeding curlew recovery initiatives by enabling early detection of threats or environmental stressors, informing targeted interventions to protect vulnerable nests and improve chick survival rates. This contribution to curlew population recovery is critical for reversing the alarming declines experienced by this species.

The results of the trial were promising, demonstrating high accuracy in curlew detection. The model effectively filtered out blank images triggered by moving vegetation, significantly reducing the workload for conservationists. Hosting the model on the Conservation AI platform further enhanced its scalability, allowing for broader deployment to monitor other species or to expand curlew conservation efforts.

This automated monitoring system also creates opportunities for broader community engagement. By encouraging individuals to deploy cameras in local environments—such as gardens, communal spaces, or workplaces—this approach could greatly expand the reach of conservation initiatives for other types of nesting birds – not just ground-nesting species. Organisations like GWCT can now leverage this scalable solution to enhance biodiversity monitoring, enabling a more comprehensive understanding of species distribution and population dynamics across the UK. Engaging citizen scientists in this way not only increases data collection capacity but also fosters public awareness and involvement in conservation efforts.

Overcoming the issues highlighted in section 4, future work will focus on refining model performance, particularly through guidelines for camera placement and classification accuracy of smaller species like curlew chicks. Enhancements in training data quality, data augmentation techniques, and hyperparameter tuning are also needed to train out misclassification such as those evident between common pheasants and curlew.

Overall, the study demonstrated excellent utility in monitoring ground-nesting birds. The approach provides a scalable and effective tool to support curlew conservation and recovery efforts, aligning with both immediate and long-term biodiversity goals. By making the model publicly available on the Conservation AI platform, this research empowers researchers, conservationists, and citizen scientists to contribute meaningfully to curlew population recovery and broader conservation initiatives.

This work had provided us with a solid platform for the longitudinal curlew nesting season survey in 2025 to fully evaluate the efficiency of the results produced and to understand the potential impact of this approach on curlew populations.

**Acknowledgments:** The authors would like to thank ReoLink for donating the cameras and solar panels used in this study and Vodafone UK for sponsoring our communication. Finally, the authors would like to thank Rachel Chalmers for tagging all the data used in this study and for contributing so much to the Conservation AI platform.